\documentclass{article}

\usepackage[final,sglblindworkshop]{neurips_2026}
\workshoptitle{AI Foundations for Power Grids}

\usepackage[utf8]{inputenc}
\usepackage[T1]{fontenc}
\usepackage{amsmath,amssymb}
\usepackage{booktabs}
\usepackage{multirow}
\usepackage{graphicx}
\usepackage{xcolor}
\usepackage{microtype}
\usepackage{url}
\usepackage{algorithm}
\usepackage{algpseudocode}
\usepackage{acronym}
\usepackage{float}
\usepackage[hidelinks]{hyperref}

\graphicspath{{figures/}}

\title{The Cost of Coverage: Conformalising Native\\Forecast Uncertainty for GB Wind Power}

\author{Carl Somers \\ University College London \\ \texttt{carl.somers.24@ucl.ac.uk}}

\begin{document}

\maketitle

\begin{abstract}
Uncertainty bands on wind forecasts help inform reserve sizing and commitments, but often fail to deliver their stated coverage. We apply conformal prediction to native bands to see what it adds and what it costs, comparing six forecasters, from ARIMA to a tabular foundation model, at two wind farms in Great Britain (GB) under operational conditions. We show that calibration repairs empirical marginal coverage with little or no Winkler Score penalty, but this apparent success conceals under-coverage in high winds. Calibrating by forecast wind regime turns marginal coverage into regime-level coverage, cutting day-ahead out-of-band volume in high winds by over 40\% relative to pooled calibration while narrowing bands overall.
\end{abstract}

\section{Introduction}

Wind became Great Britain’s largest source of electricity in 2024, supplying 30\% of generation and marking a major milestone in the shift from fossil fuels to renewable power \citep{neso_2024_review}. At the site level, generation volume is settled every half-hour against contracted positions with deviations cashed out at the balancing price, which rises sharply when the system is short. On the 8th of January 2025, a shortfall in forecast wind helped push prices to \pounds2,900/MWh and contributed to a single-day balancing cost of \pounds21M \citep{neso_margins_8jan}. Prediction intervals are central to decision making under this weather uncertainty, but remain useful only if they achieve their stated coverage without becoming too wide.

Modern machine learning forecasting pipelines output native prediction intervals, yet these are often miscalibrated and routinely under- or over-cover nominal targets. Parametric and Bayesian methods rely on distributional assumptions, while quantile regression depends on fit. Conformal Prediction (CP) \citep{vovk2005algorithmic} offers finite-sample, distribution-free marginal coverage (averaged over all forecasts) around an arbitrary base forecaster. This guarantee rests on exchangeability, an assumption wind violates through autocorrelation and seasonal non-stationarity. The literature responds through re-weighting \citep{barber2023conformal}, residual modelling \citep{xu2023conformal}, and online adaptation \citep{gibbs2021adaptive}.

Conformal methods have reached renewables, including conformalised quantile regression for wind \citep{jonkers2024novel}, conformal predictive distributions for wind speed \citep{althoff2023evaluation}, context-aware weighted calibration of renewable forecasts \citep{moradi2025enhanced}, and in solar nowcasting, a comparative benchmark of CP techniques over wrapped point forecasts \citep{NAM2025100573}. We ask what conformal calibration adds to the native uncertainty band a forecaster already publishes, and what it costs under realistic operational conditions.

We contribute three things: (i) a native-versus-wrapped measurement, scoring six forecaster classes against their conformalised counterparts on the same held-out sample in coverage, width and Winkler Score; (ii) an open operational benchmark with two wind farms, across three horizons, purged refits and weather data delays, with the full pipeline released; (iii) an account of how missed energy varies with wind regime and direction.

\section{Benchmark}

\paragraph{Data.} We use open supervisory control and data acquisition (SCADA) data from two operational GB farms, Kelmarsh (6 turbines, 12.3 MW) and Penmanshiel (14 turbines, 28.7 MW) at 30-minute settlement-period resolution \citep{plumley_2025_16807304, plumley_2025_16807551}. The target is the farm capacity factor (CF), output as a fraction of the reporting turbines' rated capacity, formed where at least half the farm reports, excluding stopped, curtailed or de-rated turbine observations. Numerical Weather Prediction (NWP) comes from the 31-member NOAA GEFS at 0.25 degrees ($20$km) via dynamical.org \citep{noaa_gefs_dynamical}, with wind speed at the 80m level closest to hub height. The ensemble mean drives the point forecast and the spread is kept as a feature. Realised weather for retrospective audit comes from ERA5 re-analysis \citep{hersbach2020ERA5}.

\paragraph{Training regime.} Models are refit on an expanding window with at least one year of training context, every 30 days. We evaluate three horizons ($h \in \{30\text{min}, 3\text{h}, 24\text{h}\}$) targeting 90\% out-of-sample coverage, spanning near-delivery balancing, intraday adjustments and day-ahead commitments. To replicate operational conditions we: (i) purged expanding refits by dropping training rows within one horizon of the window boundary, whose targets were not yet observed; (ii) added a 6-hour publication lag on GEFS forecasts to account for dissemination delays; and (iii) held each conformity score in a queue until its outcome is known, preventing leakage.

\paragraph{Forecasting Models.} The six models produce native bands by different mechanisms (Table~\ref{tab:subjects}), with persistence as a baseline. Appendix~\ref{app:rmse} reports point-forecast root mean squared error (RMSE).

\begin{table}[htbp]
\caption{The six native-UQ classes and the mechanism each uses to produce its own band.}
\label{tab:subjects}
\centering
\small
\setlength{\tabcolsep}{5.5pt}
\begin{tabular}{@{}llll@{}}
\toprule
Forecaster & Reference(s) & Class & Native Uncertainty Band \\
\midrule
Persistence & - & Empirical residual & Expanding-window residual quantiles \\
ARIMA & \citet{box2015time} & Parametric & State-space Gaussian forecast interval \\
LightGBM-QR & \citet{ke2017lightgbm} & Direct quantile & Pinball-optimised quantile interval \\
Bootstrap ensemble & \citet{breiman1996bagging} & Ensemble spread & Out-of-bag residual quantiles \\
TabPFN v3 & \citet{grinsztajn2026tabpfn} & Amortised Bayesian & Posterior predictive distribution \\
GEFS power curve & \citet{gneiting2005calibrated} & NWP ensemble & Censored EMOS over member scatter \\
\bottomrule
\end{tabular}
\end{table}

\paragraph{Conformal Calibration.} Conformal prediction can be used to correct an interval using its recent errors on a calibration set. We adjust the native band $[\ell_t, u_t]$ by taking the Conformalised Quantile Regression (CQR) score \citep{romano2019conformalized} for observation $y_t$,
\[ s_t = \max(\ell_t - y_t,\; y_t - u_t). \]
The corrected band becomes $[\ell_t - Q_t, u_t + Q_t]$, where $Q_t$ is the $\lceil (n+1)(1-\alpha)\rceil/n$ quantile of the $n \le 1000$ latest observed scores. This Online Sequential Split Conformal Prediction (OSSCP) guarantees coverage only under exchangeability \citep{lei2018distribution}, so for wind we evaluate coverage empirically \citep{oliveira2024split}. We also test Adaptive Conformal Inference (ACI), which dynamically adjusts the target miscoverage rate ($\alpha$) online as
\[ \alpha_{t+1} = \alpha_t +\gamma(\alpha - \mathbb{1}\{y_t \notin \hat C_t\}), \] 
with step size $\gamma$ \citep{gibbs2021adaptive}, and $\hat{C}_t$ the estimated interval at time $t$. We test OSSCP with Mondrian binning, whereby we bucket calibration by forecasted wind regime (0-5, 5-11, 11+ m/s) to improve conditional coverage, the coverage within each regime \citep{vovk2005algorithmic}. We also evaluate Aggregated ACI (AgACI) \citep{zaffran2022adaptive}, Dynamically Tuned ACI (DtACI) \citep{gibbs2024conformal} and NexCP \citep{barber2023conformal}.

\paragraph{Metrics.} We measure coverage (\% of predictive intervals containing the observed value) and mean width of those intervals. The Winkler Score \citep{winkler1972decision} \[ W_t = (u_t - \ell_t) + \frac{2}{\alpha}[(\ell_t - y_t)_+ + (y_t - u_t)_+], \]jointly penalises width and misses in CF units, so lower is better. The \emph{cost of coverage} is the change in width and Winkler Score that wrapping causes. We also express misses as the energy outside the band, $q_t = 0.5 \cdot C\,[(\ell_t - y_t)_+ + (y_t - u_t)_+]$ MWh per settlement period, with fixed installed capacity $C$ (MW), split into over- ($y_t > u_t$) and under-production ($y_t < \ell_t$).

\section{Results}
Kelmarsh is evaluated over 27-01-2024 to 30-12-2024 and Penmanshiel from 20-03-2023 to 31-12-2023. Full scorecards are in Appendix~\ref{app:scorecards}.

\begin{figure}[H]
\centering
\includegraphics[width=\linewidth]{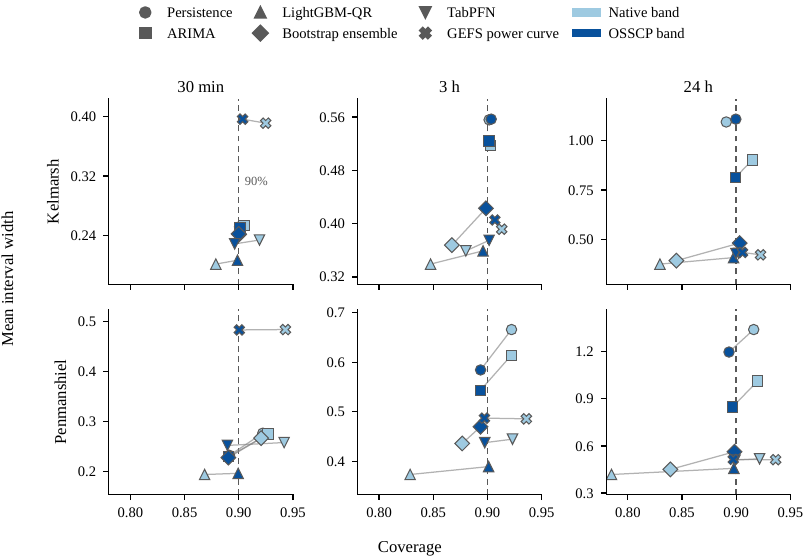}
\caption{The width cost of restoring marginal coverage. Each marker is one forecaster; the grey segment runs from its native band (light) to its OSSCP band (dark), at three horizons and both farms.}
\label{fig:width}
\end{figure}

\paragraph{Marginal coverage is repaired.} Conformal calibration brings native coverage close to the 90\% target across both farms and all horizons (Figure~\ref{fig:width}). Native bands near the target change little, while overly wide bands can narrow. At 24 hours, LightGBM-QR and the bootstrap ensemble require roughly 10–25\% wider bands to correct under-coverage. Winkler Scores are maintained or improved in 30 of 36 comparisons, with increases below 0.3\% in the remainder.

\begin{figure}[H]
\centering
\includegraphics[width=\linewidth]{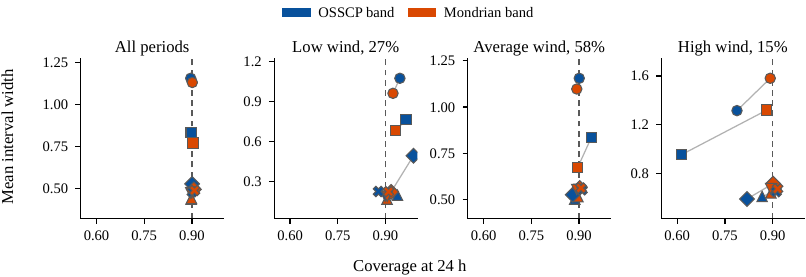}
\caption{OSSCP versus Mondrian-OSSCP day-ahead coverage and width (CF), both farms pooled, by \emph{forecast} wind regime. Percentages show regime frequency. Markers follow Figure~\ref{fig:width}.}
\label{fig:wind}
\end{figure}

\paragraph{Adaptive gains diminish with forecast horizon.} Selecting the best rule in hindsight for each farm, model, and horizon improves OSSCP's Winkler Score by 3.5\% on average at 30 minutes (up to 8.6\%), but this advantage narrows to 1.2\% at 3 hours and reverses to $-1.0\%$ at 24 hours (Appendix~\ref{app:scorecards}).

\begin{figure}[H]
\centering
\includegraphics[width=\linewidth]{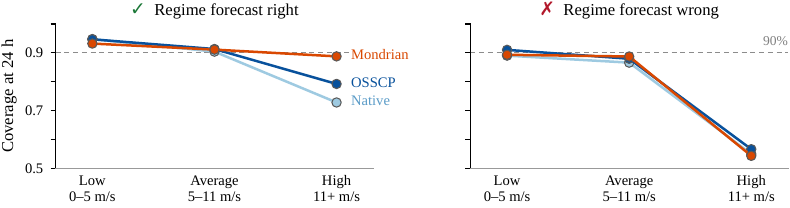}
\caption{Day-ahead coverage by \textit{realised} wind regime, both farms and models pooled, split by whether the GEFS regime forecast matched ERA5.}
\label{fig:regime}
\end{figure}

\paragraph{Average coverage hides the regimes.} At 24\,h, every OSSCP band covers close to the 90\% target overall, but by wind regime several models over-cover in low winds, offsetting under-coverage in high winds (Figure~\ref{fig:wind}). Mondrian binning brings every model to 87-92\% in forecast high winds, with mean width overall 2-4\% narrower than OSSCP's. Operationally, however, Mondrian can only bin by the \emph{forecast} regime. An audit against ERA5 \emph{realised} wind speed \citep{hersbach2020ERA5} shows that in the 16\% of realised high-wind periods that GEFS forecast as average wind, coverage falls to about 0.55 for native, OSSCP and Mondrian bands alike (Figure~\ref{fig:regime}). High-wind periods are rare but account for over a third of OSSCP's missed energy, almost all above the upper bound. Across all high-wind periods, Mondrian cuts missed energy by over 40\% relative to OSSCP.

\section{Discussion}

\paragraph{Practical Guidelines.} For day-ahead forecasts, calibrating within forecast wind regimes (Mondrian-OSSCP) is a sensible default, restoring the high-wind coverage hidden by the pooled score, with narrower bands on average. Adaptive rules are worth considering for online or short-horizon forecasts. Native bands and those calibrated using pooled or Mondrian methods all show substantial under-coverage when the weather forecast misclassifies high winds, so generators should check coverage against both forecast and realised conditions rather than relying on the headline score alone. Calibration also adds little when native bands are already reliable, as seen for TabPFN across the day-ahead forecast regimes at both farms.

\paragraph{Scoring Rules vs. Grid Reality.} The Winkler Score treats equal-sized misses alike, although their financial consequences depend on direction and system conditions. Our high-wind misses are mostly above the upper bound (over-production). If this output is uncontracted and the system is also in surplus, it may earn little or even incur charges at negative imbalance prices. Smaller shortfalls during scarcity can be more expensive, as illustrated by the high imbalance prices on the 8th of January 2025. Our missed energy metric captures volume alone, while the financial value of reliable bands requires evaluating the commitment decisions and returns they produce within a trading optimiser.

\paragraph{Limitations \& Extensions.} We study two onshore GB farms, so generalisation to offshore sites, other markets and other years remains open. Curtailed periods are excluded, so the target is not the settled metered output. As curtailment is most common in high winds, this motivates a specific classification layer for it. Realised regimes come from ERA5 reanalysis rather than on-site measurements. Extensions include multi-step joint calibration \citep{wang2024online}, context weighting \citep{moradi2025enhanced} and soft regime probabilities. AI weather forecasts \citep{bonev2025fourcastnet,weyn2026aurora, alet2026operational} could also be compared with GEFS to test whether improved regime prediction reduces high-wind under-coverage.

\clearpage
\section*{Domain Checklist}

\begin{enumerate}
\item \textbf{Physics feasibility.} AC/DC power flow not applicable as we forecast generation rather than network states or power flows. Though our capacity factors can fall slightly outside [0, 1] due to parasitic draw or over-rating, we retain these observations.
\item \textbf{Out-of-distribution evaluation.} Each pipeline is fit on its own farm, so the second farm is a replication with a different region, layout and evaluation year rather than a transfer experiment (see Figure~\ref{fig:farms}). All evaluations are run out-of-sample, and the misclassified-regime audit (Figure~\ref{fig:regime}) measures coverage when the forecast wind regime is wrong.

\begin{figure}[h]
\centering
\includegraphics[width=0.8\linewidth]{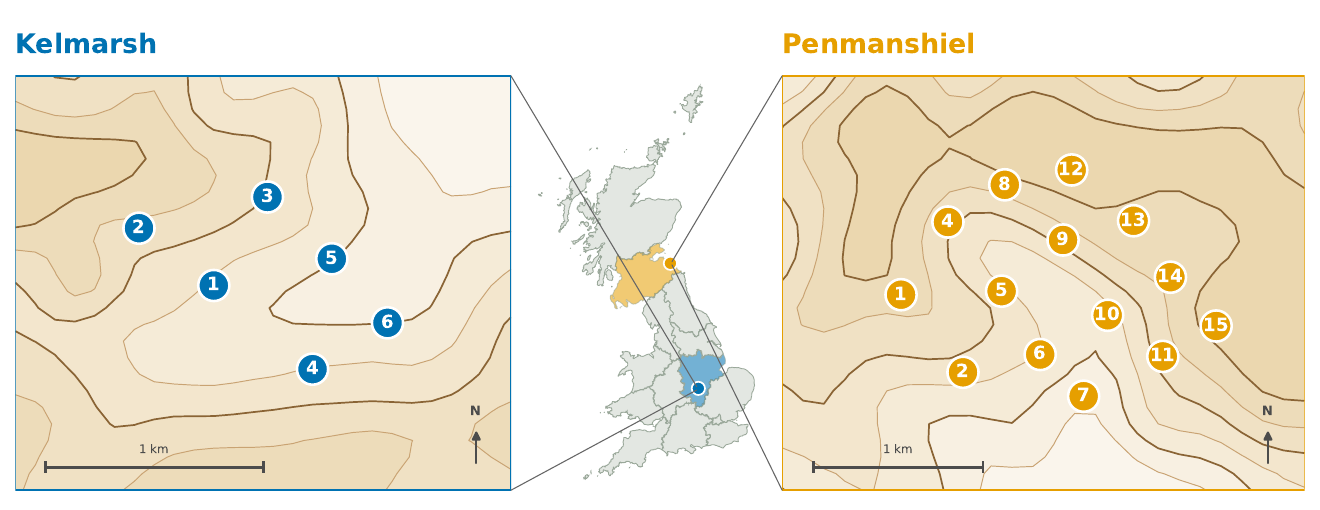}
\caption{Farm layouts and corresponding Grid Supply Point (GSP) groups. Kelmarsh: 6 turbines, 12.3 MW, East Midlands, evaluated in 2024. Penmanshiel: 14 turbines, 28.7 MW, South of Scotland, evaluated in 2023.}
\label{fig:farms}
\end{figure}
\end{enumerate}

\begin{enumerate}
\setcounter{enumi}{2}
\item \textbf{Failure modes.} Coverage falls as low as 0.55 in high winds when the forecast regime is wrong, a failure hidden by common aggregate coverage diagnostics.
\item \textbf{System-level effect.} Reserve requirements depend on forecast uncertainty. Measuring each miss by volume and direction shows the hardest regime to cover (high wind) misses mostly on the upside and has the largest mean missed energy per forecast - conditional calibration cuts that energy by over 40\%. However, system-level effects are not evaluated directly. We report missed energy by volume and direction as operational diagnostics, without modelling reserve requirements or trading decisions.
\item \textbf{Data and code.} Penmanshiel \citep{plumley_2025_16807304} and Kelmarsh \citep{plumley_2025_16807551} SCADA are open under CC-BY-4.0. NOAA GEFS via dynamical.org's public Zarr store \citep{noaa_gefs_dynamical}. All pipeline code \& Marimo notebooks open at \url{https://github.com/CSomers3/uq4wind}.
\end{enumerate}

\section*{Acknowledgments} 
The author thanks Dr F Javier Rubio (UCL) for introducing him to conformal prediction and for draft feedback.

\section*{Generative-AI / LLM Use Statement} 
LLMs assisted in coding and reviewing the pipeline evaluation and in reviewing drafts of the paper.

\newpage
\bibliographystyle{plainnat}
\bibliography{ref}

\begin{thebibliography}{29}
\providecommand{\natexlab}[1]{#1}
\providecommand{\url}[1]{\texttt{#1}}
\expandafter\ifx\csname urlstyle\endcsname\relax
  \providecommand{\doi}[1]{doi: #1}\else
  \providecommand{\doi}{doi: \begingroup \urlstyle{rm}\Url}\fi

\bibitem[Alet et~al.(2026)Alet, Andersson, Price, Markou, El-Kadi, Masters, Li, Merchant, Williams, Thornton, MacKay, Graham, Uddin, Gaiarin, Shah, Kruse, Hogsett, Zelinsky, Cangialosi, and Battaglia]{alet2026operational}
Ferran Alet, Tom Andersson, Ilan Price, Stratis Markou, Andrew El-Kadi, Dominic Masters, Amy Li, Samier Merchant, Natalie Williams, Gregory Thornton, Ken MacKay, Olivia Graham, Akib Uddin, Ben Gaiarin, Devaja Shah, Elinor Kruse, Wallace Hogsett, David Zelinsky, John Cangialosi, and Peter Battaglia.
\newblock Operational tropical cyclone forecasting with {AI}.
\newblock \emph{Nature}, 08 2026.
\newblock \doi{10.1038/s41586-026-10953-2}.

\bibitem[Althoff et~al.(2023)Althoff, Szabadváry, Anderson, and Carlsson]{althoff2023evaluation}
Simon Althoff, Johan~Hallberg Szabadváry, Jonathan Anderson, and Lars Carlsson.
\newblock Evaluation of conformal-based probabilistic forecasting methods for short-term wind speed forecasting.
\newblock In \emph{Conformal and Probabilistic Prediction with Applications}, pages 100--115. PMLR, 2023.

\bibitem[Barber et~al.(2023)Barber, Candes, Ramdas, and Tibshirani]{barber2023conformal}
Rina~Foygel Barber, Emmanuel~J Candes, Aaditya Ramdas, and Ryan~J Tibshirani.
\newblock Conformal prediction beyond exchangeability.
\newblock \emph{The Annals of Statistics}, 51\penalty0 (2):\penalty0 816--845, 2023.

\bibitem[Bonev et~al.(2025)Bonev, Kurth, Mahesh, Bisson, Kossaifi, Kashinath, Anandkumar, Collins, Pritchard, and Keller]{bonev2025fourcastnet}
Boris Bonev, Thorsten Kurth, Ankur Mahesh, Mauro Bisson, Jean Kossaifi, Karthik Kashinath, Anima Anandkumar, William~D Collins, Michael~S Pritchard, and Alexander Keller.
\newblock {FourCastNet} 3: A geometric approach to probabilistic machine-learning weather forecasting at scale.
\newblock \emph{arXiv preprint arXiv:2507.12144}, 2025.

\bibitem[Box et~al.(2015)Box, Jenkins, Reinsel, and Ljung]{box2015time}
George~EP Box, Gwilym~M Jenkins, Gregory~C Reinsel, and Greta~M Ljung.
\newblock \emph{Time series analysis: forecasting and control}.
\newblock John Wiley \& Sons, 2015.

\bibitem[Breiman(1996)]{breiman1996bagging}
Leo Breiman.
\newblock Bagging predictors.
\newblock \emph{Machine learning}, 24\penalty0 (2):\penalty0 123--140, 1996.

\bibitem[Gibbs and Candes(2021)]{gibbs2021adaptive}
Isaac Gibbs and Emmanuel Candes.
\newblock Adaptive conformal inference under distribution shift.
\newblock \emph{Advances in Neural Information Processing Systems}, 34:\penalty0 1660--1672, 2021.

\bibitem[Gibbs and Cand{\`e}s(2024)]{gibbs2024conformal}
Isaac Gibbs and Emmanuel~J Cand{\`e}s.
\newblock Conformal inference for online prediction with arbitrary distribution shifts.
\newblock \emph{Journal of Machine Learning Research}, 25\penalty0 (162):\penalty0 1--36, 2024.

\bibitem[Gneiting et~al.(2005)Gneiting, Raftery, Westveld~III, and Goldman]{gneiting2005calibrated}
Tilmann Gneiting, Adrian~E Raftery, Anton~H Westveld~III, and Tom Goldman.
\newblock Calibrated probabilistic forecasting using ensemble model output statistics and minimum {CRPS} estimation.
\newblock \emph{Monthly weather review}, 133\penalty0 (5):\penalty0 1098--1118, 2005.

\bibitem[Grinsztajn et~al.(2026)Grinsztajn, Flöge, Key, Birkel, Jund, Roof, Manium, Hoo, Bühler, Garg, Safaric, Robertson, Jäger, Alessi, Hayler, Moroshan, Purucker, Singer, and Hutter]{grinsztajn2026tabpfn}
Léo Grinsztajn, Klemens Flöge, Oscar Key, Felix Birkel, Philipp Jund, Brendan Roof, Mihir Manium, Shi~Bin Hoo, Magnus Bühler, Anurag Garg, Dominik Safaric, Jake Robertson, Benjamin Jäger, Simone Alessi, Adrian Hayler, Vladyslav Moroshan, Lennart Purucker, Philipp Singer, and Frank Hutter.
\newblock {TabPFN}-3: Technical report, 05 2026.

\bibitem[Hersbach et~al.(2020)Hersbach, Bell, Berrisford, Hirahara, Horányi, Muñoz-Sabater, Nicolas, Peubey, Radu, Schepers, Simmons, Soci, Abdalla, Abellan, Balsamo, Bechtold, Biavati, Bidlot, Bonavita, De~Chiara, Dahlgren, Dee, Diamantakis, Dragani, Flemming, Forbes, Fuentes, Geer, Haimberger, Healy, Hogan, Hólm, Janisková, Keeley, Laloyaux, Lopez, Lupu, Radnoti, de~Rosnay, Rozum, Vamborg, Villaume, and Thépaut]{hersbach2020ERA5}
Hans Hersbach, Bill Bell, Paul Berrisford, Shoji Hirahara, András Horányi, Joaquín Muñoz-Sabater, Julien Nicolas, Carole Peubey, Raluca Radu, Dinand Schepers, Adrian Simmons, Cornel Soci, Saleh Abdalla, Xavier Abellan, Gianpaolo Balsamo, Peter Bechtold, Gionata Biavati, Jean Bidlot, Massimo Bonavita, Giovanna De~Chiara, Per Dahlgren, Dick Dee, Michail Diamantakis, Rossana Dragani, Johannes Flemming, Richard Forbes, Manuel Fuentes, Alan Geer, Leo Haimberger, Sean Healy, Robin~J. Hogan, Elías Hólm, Marta Janisková, Sarah Keeley, Patrick Laloyaux, Philippe Lopez, Cristina Lupu, Gabor Radnoti, Patricia de~Rosnay, Iryna Rozum, Freja Vamborg, Sebastien Villaume, and Jean-Noël Thépaut.
\newblock The {ERA5} global reanalysis.
\newblock \emph{Quarterly Journal of the Royal Meteorological Society}, 146\penalty0 (730):\penalty0 1999--2049, 2020.
\newblock \doi{https://doi.org/10.1002/qj.3803}.
\newblock URL \url{https://rmets.onlinelibrary.wiley.com/doi/abs/10.1002/qj.3803}.

\bibitem[Jonkers et~al.(2024)Jonkers, Avendano, Van~Wallendael, and Van~Hoecke]{jonkers2024novel}
Jef Jonkers, Diego~Nieves Avendano, Glenn Van~Wallendael, and Sofie Van~Hoecke.
\newblock A novel day-ahead regional and probabilistic wind power forecasting framework using deep {CNN}s and conformalized regression forests.
\newblock \emph{Applied Energy}, 361:\penalty0 122900, 2024.

\bibitem[Ke et~al.(2017)Ke, Meng, Finley, Wang, Chen, Ma, Ye, and Liu]{ke2017lightgbm}
Guolin Ke, Qi~Meng, Thomas Finley, Taifeng Wang, Wei Chen, Weidong Ma, Qiwei Ye, and Tie-Yan Liu.
\newblock {LightGBM}: A highly efficient gradient boosting decision tree.
\newblock \emph{Advances in neural information processing systems}, 30, 2017.

\bibitem[Lei et~al.(2018)Lei, G’Sell, Rinaldo, Tibshirani, and Wasserman]{lei2018distribution}
Jing Lei, Max G’Sell, Alessandro Rinaldo, Ryan~J Tibshirani, and Larry Wasserman.
\newblock Distribution-free predictive inference for regression.
\newblock \emph{Journal of the American Statistical Association}, 113\penalty0 (523):\penalty0 1094--1111, 2018.

\bibitem[Moradi et~al.(2025)Moradi, Tanneau, Zandehshahvar, and Van~Hentenryck]{moradi2025enhanced}
Alireza Moradi, Mathieu Tanneau, Reza Zandehshahvar, and Pascal Van~Hentenryck.
\newblock Enhanced renewable energy forecasting using context-aware conformal prediction.
\newblock \emph{arXiv preprint arXiv:2510.15780}, 2025.

\bibitem[Nam et~al.(2025)Nam, Ogliari, Leva, Pafumi, Alberti, and Duong]{NAM2025100573}
Nguyen~Binh Nam, Emanuele Ogliari, Sonia Leva, Emilio Pafumi, Davide Alberti, and Minh~Quan Duong.
\newblock Comparative analysis of conformal prediction techniques and machine learning models for very short-term solar power forecasting.
\newblock \emph{Energy and AI}, 21:\penalty0 100573, 2025.
\newblock ISSN 2666-5468.
\newblock \doi{https://doi.org/10.1016/j.egyai.2025.100573}.
\newblock URL \url{https://www.sciencedirect.com/science/article/pii/S2666546825001053}.

\bibitem[{National Energy System Operator (NESO)}(2025{\natexlab{a}})]{neso_2024_review}
{National Energy System Operator (NESO)}.
\newblock Britain's electricity explained: 2024 review, January 2025{\natexlab{a}}.
\newblock URL \url{https://www.neso.energy/news/britains-electricity-explained-2024-review}.

\bibitem[{National Energy System Operator (NESO)}(2025{\natexlab{b}})]{neso_margins_8jan}
{National Energy System Operator (NESO)}.
\newblock What happened to margins on 8 january.
\newblock https://www.neso.energy/news/what-happened-margins-8-january, 2025{\natexlab{b}}.
\newblock Accessed: 2026-08-29.

\bibitem[{NOAA Global Ensemble Forecast System (GEFS)}(2026)]{noaa_gefs_dynamical}
{NOAA Global Ensemble Forecast System (GEFS)}.
\newblock {NOAA GEFS} - dynamical.org icechunk zarr, 2026.
\newblock URL \url{https://registry.opendata.aws/dynamical-noaa-gefs}.

\bibitem[Oliveira et~al.(2024)Oliveira, Orenstein, Ramos, and Romano]{oliveira2024split}
Roberto~I Oliveira, Paulo Orenstein, Thiago Ramos, and Jo{\~a}o~Vitor Romano.
\newblock Split conformal prediction and non-exchangeable data.
\newblock \emph{Journal of Machine Learning Research}, 25\penalty0 (225):\penalty0 1--38, 2024.

\bibitem[Plumley and Takeuchi(2025{\natexlab{a}})]{plumley_2025_16807304}
Charlie Plumley and Roberta Takeuchi.
\newblock Penmanshiel wind farm data, August 2025{\natexlab{a}}.
\newblock URL \url{https://doi.org/10.5281/zenodo.16807304}.

\bibitem[Plumley and Takeuchi(2025{\natexlab{b}})]{plumley_2025_16807551}
Charlie Plumley and Roberta Takeuchi.
\newblock Kelmarsh wind farm data, August 2025{\natexlab{b}}.
\newblock URL \url{https://doi.org/10.5281/zenodo.16807551}.

\bibitem[Romano et~al.(2019)Romano, Patterson, and Candes]{romano2019conformalized}
Yaniv Romano, Evan Patterson, and Emmanuel Candes.
\newblock Conformalized quantile regression.
\newblock \emph{Advances in neural information processing systems}, 32, 2019.

\bibitem[Vovk et~al.(2005)Vovk, Gammerman, and Shafer]{vovk2005algorithmic}
Vladimir Vovk, Alexander Gammerman, and Glenn Shafer.
\newblock \emph{Algorithmic learning in a random world}, volume~29.
\newblock Springer, 2005.

\bibitem[Wang and Hyndman(2024)]{wang2024online}
Xiaoqian Wang and Rob~J Hyndman.
\newblock Online conformal inference for multi-step time series forecasting.
\newblock \emph{arXiv preprint arXiv:2410.13115}, 2024.

\bibitem[Weyn et~al.(2026)Weyn, Ni, Misra, Fein, Dong, Bruinsma, Turner, Corey, Thambiratnam, White, and Takeda]{weyn2026aurora}
Jonathan Weyn, Zekun Ni, Amit Misra, Will Fein, Haiyu Dong, Wessel~P. Bruinsma, Richard~E. Turner, Matt Corey, Kit Thambiratnam, Kevin White, and Kenji Takeda.
\newblock Aurora 1.5: Fine-tuning a foundation model for medium-range ensemble weather prediction.
\newblock July 2026.
\newblock URL \url{https://www.microsoft.com/en-us/research/publication/aurora-1-5/}.

\bibitem[Winkler(1972)]{winkler1972decision}
Robert~L Winkler.
\newblock A decision-theoretic approach to interval estimation.
\newblock \emph{Journal of the American Statistical Association}, 67\penalty0 (337):\penalty0 187--191, 1972.

\bibitem[Xu and Xie(2023)]{xu2023conformal}
Chen Xu and Yao Xie.
\newblock Conformal prediction for time series.
\newblock \emph{IEEE transactions on pattern analysis and machine intelligence}, 45\penalty0 (10):\penalty0 11575--11587, 2023.

\bibitem[Zaffran et~al.(2022)Zaffran, F{\'e}ron, Goude, Josse, and Dieuleveut]{zaffran2022adaptive}
Margaux Zaffran, Olivier F{\'e}ron, Yannig Goude, Julie Josse, and Aymeric Dieuleveut.
\newblock Adaptive conformal predictions for time series.
\newblock In \emph{International conference on machine learning}, pages 25834--25866. PMLR, 2022.

\end{thebibliography}

\clearpage
\appendix
\raggedbottom

\section{Full Scorecards}
\label{app:scorecards}

\begin{table}[H]
\caption{Full scorecard, Kelmarsh 2024 ($N$ = 5,652-5,911 per cell, target $1-\alpha = 0.90$).}
\label{tab:kelmarsh}
\centering
\resizebox{\linewidth}{!}{\begin{tabular}{@{}llccccccccc@{}}
\toprule
 &  & \multicolumn{3}{c}{Coverage} & \multicolumn{3}{c}{Width} & \multicolumn{3}{c}{Winkler Score} \\
 &  & 30 min & 3 h & 24 h & 30 min & 3 h & 24 h & 30 min & 3 h & 24 h \\
Subject & Method &  &  &  &  &  &  &  &  &  \\
\midrule
\multirow[t]{8}{*}{Persistence} & Native & 0.90 & 0.90 & 0.89 & 0.25 & 0.56 & 1.09 & 0.39 & 0.81 & 1.47 \\
 & OSSCP & 0.90 & 0.90 & 0.90 & 0.25 & 0.56 & 1.11 & 0.39 & 0.79 & 1.44 \\
 & NexCP & 0.91 & 0.90 & 0.89 & 0.26 & 0.56 & 1.09 & 0.37 & 0.78 & 1.45 \\
 & ACI ($\gamma=0.01$) & 0.90 & 0.90 & 0.90 & 0.25 & 0.55 & 1.17 & 0.38 & 0.78 & 1.48 \\
 & ACI ($\gamma=0.05$) & 0.90 & 0.90 & 0.90 & 0.26 & 0.62 & 1.29 & 0.38 & 0.83 & 1.68 \\
 & AgACI & 0.90 & 0.90 & 0.90 & 0.25 & 0.57 & 1.17 & 0.37 & 0.78 & 1.48 \\
 & DtACI & 0.90 & 0.90 & 0.88 & 0.24 & 0.54 & 1.07 & 0.37 & 0.77 & 1.50 \\
 & Mondrian-OSSCP & 0.91 & 0.91 & 0.91 & 0.24 & 0.53 & 1.12 & 0.35 & 0.73 & 1.40 \\
\cmidrule{1-11}
\multirow[t]{8}{*}{ARIMA} & Native & 0.90 & 0.90 & 0.92 & 0.25 & 0.52 & 0.90 & 0.38 & 0.75 & 1.12 \\
 & OSSCP & 0.90 & 0.90 & 0.90 & 0.25 & 0.52 & 0.81 & 0.38 & 0.73 & 1.06 \\
 & NexCP & 0.91 & 0.90 & 0.90 & 0.25 & 0.52 & 0.81 & 0.37 & 0.72 & 1.06 \\
 & ACI ($\gamma=0.01$) & 0.90 & 0.90 & 0.90 & 0.25 & 0.52 & 0.85 & 0.37 & 0.73 & 1.11 \\
 & ACI ($\gamma=0.05$) & 0.90 & 0.90 & 0.90 & 0.26 & 0.60 & 1.07 & 0.37 & 0.78 & 1.29 \\
 & AgACI & 0.90 & 0.90 & 0.90 & 0.25 & 0.53 & 0.84 & 0.37 & 0.73 & 1.09 \\
 & DtACI & 0.90 & 0.89 & 0.88 & 0.24 & 0.51 & 0.79 & 0.36 & 0.72 & 1.11 \\
 & Mondrian-OSSCP & 0.91 & 0.91 & 0.91 & 0.24 & 0.50 & 0.75 & 0.35 & 0.68 & 0.91 \\
\cmidrule{1-11}
\multirow[t]{8}{*}{LightGBM-QR} & Native & 0.88 & 0.85 & 0.83 & 0.20 & 0.34 & 0.38 & 0.28 & 0.49 & 0.56 \\
 & OSSCP & 0.90 & 0.90 & 0.90 & 0.21 & 0.36 & 0.41 & 0.28 & 0.48 & 0.55 \\
 & NexCP & 0.91 & 0.90 & 0.90 & 0.21 & 0.37 & 0.43 & 0.28 & 0.49 & 0.56 \\
 & ACI ($\gamma=0.01$) & 0.90 & 0.90 & 0.90 & 0.21 & 0.37 & 0.45 & 0.28 & 0.49 & 0.58 \\
 & ACI ($\gamma=0.05$) & 0.90 & 0.90 & 0.90 & 0.23 & 0.45 & 0.71 & 0.29 & 0.55 & 0.83 \\
 & AgACI & 0.90 & 0.90 & 0.90 & 0.21 & 0.37 & 0.44 & 0.28 & 0.49 & 0.57 \\
 & DtACI & 0.90 & 0.90 & 0.89 & 0.21 & 0.38 & 0.45 & 0.28 & 0.50 & 0.59 \\
 & Mondrian-OSSCP & 0.90 & 0.90 & 0.90 & 0.21 & 0.36 & 0.41 & 0.28 & 0.48 & 0.55 \\
\cmidrule{1-11}
\multirow[t]{8}{*}{Bootstrap ensemble} & Native & 0.90 & 0.87 & 0.84 & 0.24 & 0.37 & 0.39 & 0.37 & 0.62 & 0.71 \\
 & OSSCP & 0.90 & 0.90 & 0.90 & 0.24 & 0.42 & 0.48 & 0.37 & 0.60 & 0.67 \\
 & NexCP & 0.91 & 0.90 & 0.89 & 0.25 & 0.43 & 0.49 & 0.36 & 0.60 & 0.69 \\
 & ACI ($\gamma=0.01$) & 0.90 & 0.90 & 0.90 & 0.24 & 0.42 & 0.52 & 0.36 & 0.60 & 0.72 \\
 & ACI ($\gamma=0.05$) & 0.90 & 0.90 & 0.90 & 0.25 & 0.47 & 0.77 & 0.36 & 0.65 & 0.96 \\
 & AgACI & 0.90 & 0.90 & 0.90 & 0.24 & 0.43 & 0.51 & 0.36 & 0.60 & 0.70 \\
 & DtACI & 0.90 & 0.89 & 0.88 & 0.24 & 0.42 & 0.49 & 0.35 & 0.60 & 0.73 \\
 & Mondrian-OSSCP & 0.90 & 0.91 & 0.91 & 0.23 & 0.40 & 0.46 & 0.33 & 0.56 & 0.63 \\
\cmidrule{1-11}
\multirow[t]{8}{*}{TabPFN} & Native & 0.92 & 0.88 & 0.90 & 0.23 & 0.36 & 0.43 & 0.29 & 0.51 & 0.58 \\
 & OSSCP & 0.90 & 0.90 & 0.90 & 0.23 & 0.37 & 0.43 & 0.29 & 0.51 & 0.58 \\
 & NexCP & 0.91 & 0.90 & 0.90 & 0.23 & 0.39 & 0.45 & 0.29 & 0.52 & 0.58 \\
 & ACI ($\gamma=0.01$) & 0.90 & 0.90 & 0.90 & 0.23 & 0.39 & 0.47 & 0.30 & 0.52 & 0.60 \\
 & ACI ($\gamma=0.05$) & 0.90 & 0.90 & 0.90 & 0.25 & 0.47 & 0.78 & 0.31 & 0.59 & 0.89 \\
 & AgACI & 0.90 & 0.90 & 0.90 & 0.23 & 0.39 & 0.45 & 0.30 & 0.52 & 0.59 \\
 & DtACI & 0.90 & 0.90 & 0.89 & 0.23 & 0.39 & 0.47 & 0.30 & 0.52 & 0.61 \\
 & Mondrian-OSSCP & 0.89 & 0.90 & 0.89 & 0.23 & 0.38 & 0.43 & 0.29 & 0.51 & 0.58 \\
\cmidrule{1-11}
\multirow[t]{8}{*}{GEFS power curve} & Native & 0.92 & 0.91 & 0.92 & 0.39 & 0.39 & 0.42 & 0.51 & 0.54 & 0.56 \\
 & OSSCP & 0.90 & 0.91 & 0.91 & 0.40 & 0.41 & 0.44 & 0.51 & 0.54 & 0.56 \\
 & NexCP & 0.90 & 0.90 & 0.90 & 0.40 & 0.42 & 0.44 & 0.51 & 0.54 & 0.57 \\
 & ACI ($\gamma=0.01$) & 0.90 & 0.90 & 0.90 & 0.40 & 0.41 & 0.47 & 0.50 & 0.54 & 0.59 \\
 & ACI ($\gamma=0.05$) & 0.90 & 0.90 & 0.90 & 0.42 & 0.51 & 0.74 & 0.51 & 0.62 & 0.83 \\
 & AgACI & 0.90 & 0.90 & 0.90 & 0.41 & 0.41 & 0.45 & 0.50 & 0.54 & 0.57 \\
 & DtACI & 0.90 & 0.90 & 0.88 & 0.40 & 0.42 & 0.46 & 0.50 & 0.55 & 0.60 \\
 & Mondrian-OSSCP & 0.90 & 0.91 & 0.91 & 0.40 & 0.41 & 0.44 & 0.51 & 0.54 & 0.57 \\
\cmidrule{1-11}
\bottomrule
\end{tabular}
}
\end{table}

\begin{table}[H]
\caption{Full scorecard, Penmanshiel 2023 ($N$ = 6,668-6,746 per cell, target $1-\alpha = 0.90$).}
\label{tab:penmanshiel}
\centering
\resizebox{\linewidth}{!}{\begin{tabular}{@{}llccccccccc@{}}
\toprule
 &  & \multicolumn{3}{c}{Coverage} & \multicolumn{3}{c}{Width} & \multicolumn{3}{c}{Winkler Score} \\
 &  & 30 min & 3 h & 24 h & 30 min & 3 h & 24 h & 30 min & 3 h & 24 h \\
Subject & Method &  &  &  &  &  &  &  &  &  \\
\midrule
\multirow[t]{8}{*}{Persistence} & Native & 0.92 & 0.92 & 0.92 & 0.28 & 0.67 & 1.34 & 0.39 & 0.88 & 1.59 \\
 & OSSCP & 0.89 & 0.89 & 0.89 & 0.23 & 0.58 & 1.20 & 0.38 & 0.86 & 1.54 \\
 & NexCP & 0.90 & 0.90 & 0.89 & 0.24 & 0.59 & 1.16 & 0.36 & 0.83 & 1.52 \\
 & ACI ($\gamma=0.01$) & 0.90 & 0.90 & 0.90 & 0.23 & 0.59 & 1.27 & 0.36 & 0.83 & 1.59 \\
 & ACI ($\gamma=0.05$) & 0.90 & 0.90 & 0.90 & 0.25 & 0.65 & 1.34 & 0.36 & 0.89 & 1.69 \\
 & AgACI & 0.90 & 0.91 & 0.91 & 0.24 & 0.61 & 1.26 & 0.35 & 0.82 & 1.52 \\
 & DtACI & 0.90 & 0.89 & 0.86 & 0.23 & 0.57 & 1.13 & 0.35 & 0.83 & 1.61 \\
 & Mondrian-OSSCP & 0.90 & 0.90 & 0.89 & 0.22 & 0.55 & 1.14 & 0.34 & 0.77 & 1.48 \\
\cmidrule{1-11}
\multirow[t]{8}{*}{ARIMA} & Native & 0.93 & 0.92 & 0.92 & 0.27 & 0.61 & 1.01 & 0.38 & 0.81 & 1.20 \\
 & OSSCP & 0.89 & 0.89 & 0.90 & 0.23 & 0.54 & 0.85 & 0.37 & 0.79 & 1.12 \\
 & NexCP & 0.90 & 0.90 & 0.90 & 0.23 & 0.55 & 0.86 & 0.35 & 0.76 & 1.12 \\
 & ACI ($\gamma=0.01$) & 0.90 & 0.90 & 0.90 & 0.23 & 0.55 & 0.93 & 0.35 & 0.77 & 1.17 \\
 & ACI ($\gamma=0.05$) & 0.90 & 0.90 & 0.90 & 0.24 & 0.61 & 1.08 & 0.35 & 0.83 & 1.39 \\
 & AgACI & 0.90 & 0.90 & 0.90 & 0.23 & 0.56 & 0.90 & 0.34 & 0.77 & 1.15 \\
 & DtACI & 0.90 & 0.89 & 0.87 & 0.22 & 0.53 & 0.87 & 0.34 & 0.77 & 1.18 \\
 & Mondrian-OSSCP & 0.90 & 0.90 & 0.89 & 0.22 & 0.52 & 0.78 & 0.33 & 0.71 & 0.97 \\
\cmidrule{1-11}
\multirow[t]{8}{*}{LightGBM-QR} & Native & 0.87 & 0.83 & 0.78 & 0.19 & 0.37 & 0.42 & 0.26 & 0.51 & 0.62 \\
 & OSSCP & 0.90 & 0.90 & 0.90 & 0.20 & 0.39 & 0.46 & 0.26 & 0.51 & 0.62 \\
 & NexCP & 0.91 & 0.91 & 0.90 & 0.20 & 0.40 & 0.48 & 0.26 & 0.51 & 0.62 \\
 & ACI ($\gamma=0.01$) & 0.90 & 0.90 & 0.90 & 0.20 & 0.40 & 0.51 & 0.26 & 0.52 & 0.65 \\
 & ACI ($\gamma=0.05$) & 0.90 & 0.90 & 0.90 & 0.22 & 0.48 & 0.73 & 0.27 & 0.58 & 0.83 \\
 & AgACI & 0.90 & 0.90 & 0.90 & 0.20 & 0.40 & 0.50 & 0.26 & 0.52 & 0.64 \\
 & DtACI & 0.90 & 0.90 & 0.88 & 0.20 & 0.41 & 0.50 & 0.26 & 0.52 & 0.65 \\
 & Mondrian-OSSCP & 0.90 & 0.90 & 0.90 & 0.20 & 0.40 & 0.46 & 0.26 & 0.51 & 0.61 \\
\cmidrule{1-11}
\multirow[t]{8}{*}{Bootstrap ensemble} & Native & 0.92 & 0.88 & 0.84 & 0.27 & 0.44 & 0.45 & 0.37 & 0.70 & 0.85 \\
 & OSSCP & 0.89 & 0.89 & 0.90 & 0.23 & 0.47 & 0.56 & 0.36 & 0.69 & 0.82 \\
 & NexCP & 0.90 & 0.90 & 0.89 & 0.23 & 0.48 & 0.57 & 0.35 & 0.67 & 0.83 \\
 & ACI ($\gamma=0.01$) & 0.90 & 0.90 & 0.90 & 0.23 & 0.48 & 0.65 & 0.34 & 0.68 & 0.89 \\
 & ACI ($\gamma=0.05$) & 0.90 & 0.90 & 0.90 & 0.24 & 0.56 & 0.88 & 0.35 & 0.74 & 1.10 \\
 & AgACI & 0.91 & 0.90 & 0.90 & 0.23 & 0.49 & 0.60 & 0.34 & 0.68 & 0.84 \\
 & DtACI & 0.90 & 0.89 & 0.86 & 0.22 & 0.48 & 0.59 & 0.33 & 0.68 & 0.90 \\
 & Mondrian-OSSCP & 0.90 & 0.90 & 0.90 & 0.22 & 0.44 & 0.52 & 0.33 & 0.62 & 0.73 \\
\cmidrule{1-11}
\multirow[t]{8}{*}{TabPFN} & Native & 0.94 & 0.92 & 0.92 & 0.26 & 0.45 & 0.52 & 0.31 & 0.55 & 0.63 \\
 & OSSCP & 0.89 & 0.90 & 0.90 & 0.25 & 0.44 & 0.51 & 0.30 & 0.55 & 0.63 \\
 & NexCP & 0.91 & 0.90 & 0.90 & 0.25 & 0.45 & 0.52 & 0.30 & 0.55 & 0.64 \\
 & ACI ($\gamma=0.01$) & 0.90 & 0.90 & 0.90 & 0.25 & 0.45 & 0.55 & 0.30 & 0.55 & 0.66 \\
 & ACI ($\gamma=0.05$) & 0.90 & 0.90 & 0.90 & 0.26 & 0.56 & 0.84 & 0.31 & 0.65 & 0.92 \\
 & AgACI & 0.90 & 0.90 & 0.90 & 0.25 & 0.45 & 0.52 & 0.30 & 0.55 & 0.64 \\
 & DtACI & 0.90 & 0.90 & 0.88 & 0.25 & 0.47 & 0.55 & 0.30 & 0.57 & 0.65 \\
 & Mondrian-OSSCP & 0.89 & 0.89 & 0.90 & 0.25 & 0.44 & 0.51 & 0.30 & 0.55 & 0.63 \\
\cmidrule{1-11}
\multirow[t]{8}{*}{GEFS power curve} & Native & 0.94 & 0.94 & 0.94 & 0.48 & 0.49 & 0.51 & 0.62 & 0.63 & 0.65 \\
 & OSSCP & 0.90 & 0.90 & 0.90 & 0.48 & 0.49 & 0.51 & 0.62 & 0.63 & 0.65 \\
 & NexCP & 0.91 & 0.90 & 0.90 & 0.50 & 0.51 & 0.54 & 0.61 & 0.63 & 0.67 \\
 & ACI ($\gamma=0.01$) & 0.90 & 0.90 & 0.90 & 0.49 & 0.52 & 0.59 & 0.61 & 0.64 & 0.72 \\
 & ACI ($\gamma=0.05$) & 0.90 & 0.90 & 0.90 & 0.51 & 0.61 & 0.81 & 0.61 & 0.71 & 0.90 \\
 & AgACI & 0.91 & 0.90 & 0.90 & 0.50 & 0.51 & 0.54 & 0.60 & 0.63 & 0.67 \\
 & DtACI & 0.90 & 0.89 & 0.88 & 0.49 & 0.52 & 0.55 & 0.60 & 0.64 & 0.68 \\
 & Mondrian-OSSCP & 0.91 & 0.91 & 0.91 & 0.49 & 0.50 & 0.52 & 0.62 & 0.63 & 0.66 \\
\cmidrule{1-11}
\bottomrule
\end{tabular}
}
\end{table}

\clearpage

\section{Conditional Coverage}
\label{app:conditional}

\begin{table}[H]
\caption{Coverage by forecast wind band at 24 hours, native versus wrapped, both farms.}
\label{tab:bands}
\centering
\resizebox{\linewidth}{!}{\begin{tabular}{@{}llcccccc@{}}
\toprule
 &  & \multicolumn{2}{c}{Low (0--5 m/s)} & \multicolumn{2}{c}{Average (5--11 m/s)} & \multicolumn{2}{c}{High (11+ m/s)} \\
 &  & Kelmarsh & Penmanshiel & Kelmarsh & Penmanshiel & Kelmarsh & Penmanshiel \\
Subject & Method &  &  &  &  &  &  \\
\midrule
\multirow[t]{3}{*}{Persistence} & Native & 0.93 & 0.97 & 0.91 & 0.92 & 0.67 & 0.80 \\
 & OSSCP & 0.94 & 0.95 & 0.91 & 0.89 & 0.75 & 0.81 \\
 & Mondrian-OSSCP & 0.96 & 0.90 & 0.90 & 0.89 & 0.87 & 0.91 \\
\cmidrule{1-8}
\multirow[t]{3}{*}{ARIMA} & Native & 0.99 & 1.00 & 0.96 & 0.97 & 0.45 & 0.65 \\
 & OSSCP & 0.96 & 0.97 & 0.94 & 0.94 & 0.53 & 0.65 \\
 & Mondrian-OSSCP & 0.97 & 0.91 & 0.90 & 0.89 & 0.88 & 0.88 \\
\cmidrule{1-8}
\multirow[t]{3}{*}{LightGBM-QR} & Native & 0.82 & 0.75 & 0.84 & 0.80 & 0.81 & 0.77 \\
 & OSSCP & 0.94 & 0.94 & 0.88 & 0.89 & 0.89 & 0.86 \\
 & Mondrian-OSSCP & 0.91 & 0.90 & 0.89 & 0.90 & 0.88 & 0.90 \\
\cmidrule{1-8}
\multirow[t]{3}{*}{Bootstrap ensemble} & Native & 0.98 & 0.99 & 0.81 & 0.83 & 0.73 & 0.63 \\
 & OSSCP & 0.99 & 0.99 & 0.87 & 0.89 & 0.87 & 0.79 \\
 & Mondrian-OSSCP & 0.93 & 0.91 & 0.90 & 0.90 & 0.89 & 0.91 \\
\cmidrule{1-8}
\multirow[t]{3}{*}{TabPFN} & Native & 0.92 & 0.92 & 0.90 & 0.92 & 0.89 & 0.92 \\
 & OSSCP & 0.92 & 0.88 & 0.89 & 0.91 & 0.89 & 0.90 \\
 & Mondrian-OSSCP & 0.91 & 0.91 & 0.89 & 0.90 & 0.88 & 0.90 \\
\cmidrule{1-8}
\multirow[t]{3}{*}{GEFS power curve} & Native & 0.97 & 0.97 & 0.91 & 0.93 & 0.90 & 0.90 \\
 & OSSCP & 0.91 & 0.85 & 0.90 & 0.92 & 0.93 & 0.90 \\
 & Mondrian-OSSCP & 0.93 & 0.90 & 0.90 & 0.91 & 0.90 & 0.92 \\
\cmidrule{1-8}
\bottomrule
\end{tabular}
}
\end{table}

\section{Missed Energy}
\label{app:exposure}

\begin{table}[H]

\caption{Day-ahead \textit{missed energy} outside the uncertainty bands during realised high winds
($\geq 11$\,m/s), pooled over six models. MWh is the mean per high-wind
forecast, including zeros for covered periods.}
\label{tab:exposure}
\centering
\small
\begin{tabular}{@{}llrrr@{}}
\toprule
Farm & Method
& \shortstack{Missed energy\\(MWh)}
& \shortstack{Share of total\\missed energy (\%)}
& \shortstack{Above upper\\bound (\%)} \\
\midrule
\multirow[t]{3}{*}{Kelmarsh}
 & Native         & 0.26 & 34 & 94 \\
 & OSSCP          & 0.23 & 35 & 94 \\
 & Mondrian-OSSCP & 0.13 & 23 & 90 \\
\midrule
\multirow[t]{3}{*}{Penmanshiel}
 & Native         & 0.50 & 38 & 91 \\
 & OSSCP          & 0.50 & 38 & 93 \\
 & Mondrian-OSSCP & 0.29 & 25 & 92 \\
\bottomrule
\end{tabular}
% \resizebox{\linewidth}{!}{\results{table_missed_energy.tex}}
\end{table}

\section{Point Accuracy}
\label{app:rmse}

\begin{figure}[H]
\centering
\includegraphics[width=\linewidth]{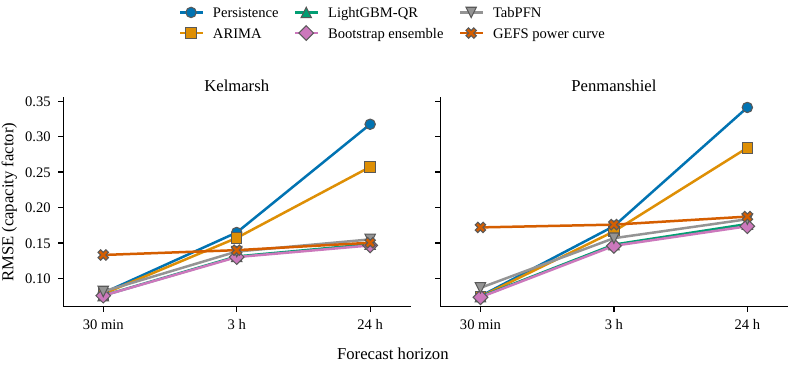}
\caption{Out-of-sample RMSE of each forecaster's point forecast (its predictive mean) at each horizon, over 2024 for Kelmarsh and 2023 for Penmanshiel, in capacity-factor units.}
\label{fig:rmse}
\end{figure}

\end{document}